\documentclass[11pt,letterpaper]{article}
\usepackage{emnlp2017}
\usepackage{times}
\usepackage{latexsym}

\emnlpfinalcopy

\def\emnlppaperid{***}

\title{Hypothesis Testing based Intrinsic Evaluation of Word Embeddings}

\author{Nishant Gurnani \\
  Department of Mathematics \\
  University of California San Diego\\
  {\tt ndgurnan@ucsd.edu}
}

\begin{document}

\maketitle

\begin{abstract}
We introduce the cross-match test - an exact, distribution free, high-dimensional hypothesis test as an intrinsic evaluation metric for word embeddings. We show that cross-match is an effective means of measuring distributional similarity between different vector representations and of evaluating the statistical significance of different vector embedding models. Additionally, we find that cross-match can be used to provide a quantitative measure of linguistic similarity for selecting bridge languages for machine translation. We demonstrate that the results of the hypothesis test align with our expectations and note that the framework of two sample hypothesis testing is not limited to word embeddings and can be extended to all vector representations.
\end{abstract}

\section{Introduction}

% Introduce problem
Word embeddings obtained via specialized models \cite{Brown1992, Pennington2014, Mikolov2013a} or neural networks \cite{Bengio2003} have been successfully used to address various natural language processing tasks \cite{Vaswani2013, Soricut2015}. These embeddings provide a nuanced representation of words that can capture various syntactic and semantic properties of natural language \cite{Mikolov2013b}. Despite their effectiveness in downstream applications, embeddings have limited practical value as standalone items. Consequently, an intrinsic evaluation metric must provide insight on the downstream task the embeddings are designed for. In this work, we use Cross-match \cite{Rosenbaum2005} - an exact, distribution free, high-dimensional hypothesis test to propose a novel approach for intrinsic evaluation of word embeddings, one that provides insight on tasks that depend on linguistic similarity.

% Evaluation is difficult
Evaluating general purpose vector representations is difficult. They are trained using simple objectives and applied to a variety of downstream tasks, thus making no single extrinsic evaluation definitive. Often, due to computational constraints, direct downstream evaluations are also impractical. In the case of word embeddings, these constraints have led to the development of dedicated evaluation tasks like similarity and analogy \cite{Rohde2006, Levy2015} which are not directly related to training objectives or to downstream tasks. Despite their ease of interpretability, \citet{Faruqui2016} have shown that these tasks do not correlate well with downstream performance. In related work, \citet{Tsvetkov2016} propose an evaluation measure QVEC-CCA that is shown to correlate well with downstream semantic tasks where the objective is to quantify the linguistic content of word embeddings by maximizing the correlation with a manually annotated linguistic resource.

In this work, we use the Cross-match hypothesis test \cite{Rosenbaum2005} to measure distributional similarity between different word vector representations. Cross-match is an adjacency based test traditionally used in clinical settings where the goal is to assess no treatment effect on a high-dimensional outcome in a randomized experiment. In our setting, we assume there exists some unknown distribution $W$ from which our constructed word embeddings $\{\mathbf{w_1,\dots,w_n}\}$ are ``sampled" from. Given two sets of word embeddings, cross-match tests whether the underlying distribution from which the embeddings were ``sampled" are identical or not. The test uses optimal non-bipartite matching to pair vectors from both sets of embeddings based on distance (e.g. a vector will be paired with it's nearest neighbor based on some distance metric). The cross-match test statistic $C$ is the number of times that a vector from one set is paired with a vector from another. The null hypothesis assumes that the vectors were sampled from the same distribution and rejects for small values of $C$. Thus, a large number of cross-matches between two sets of word embeddings suggests that they are from the same embedding distribution.

Using cross-match, we propose two illustrative examples of intrinsic evaluation. First, we use pre-trained word vectors (trained on Wikipedia using the skip-gram model in \citet{Bojanowski2016}) from Facebook's fastText library for several languages to calculate the cross-match statistic for several language pairs. We hypothesize that for linguistically similar languages, a larger statistic will be observed. Secondly, we use cross-match to assess the statistical significant of word embedding models. We consider several well known models trained on the same corpus and use cross-match to assess whether the respective word vector representations are statistically significantly different.  We hypothesize that the number of cross-matches between two different embedding models is small, thus suggesting that they capture fundamentally different linguistic aspects of the corpus.

% Organization
This paper is organized as follows: Section 2 introduces the cross-match test in detail. Experiments on embedding similarity and evaluation are described in Section 3. We discuss extensions and conclude in Section 4. 

\section{Cross-Match Test}

The cross-match test \cite{Rosenbaum2005} is a nonparametric goodness-of-fit test in arbitrary dimensions. It is an exact, distribution-free, two-sample hypothesis test that measures whether two distributions are equal or not. Formally, given two independent samples ${w_1,\dots,w_n} \sim W$ and ${v_1,\dots,v_m} \sim V$, cross-match tests the null hypothesis $H_0: W = V$ versus the alternative hypothesis $H_1: W \neq V$. The test has been traditionally used in clinical settings, where the goal is to assess no treatment effect on a high-dimensional outcome between control and treated subjects in a randomized experiment \cite{Heller2010}. In the case of word embeddings, the goal is to test whether two sets of word embedding vectors have been ``sampled" from the same distribution.

\subsection{Definition of the Cross-Match Statistic}

Let $W,V$ denote two word embedding distributions (distributions of word embedding vectors over a corpus), suppose we obtain two sets of word vectors $\{\mathbf{w_1,\dots,w_n}\} \sim W$ and $\{\mathbf{v_1,\dots,v_m}\} \sim V$. Assign the group labels 0 and 1 to indicate which sample the vectors are from such that the data are organized as follows: $\{(0,\mathbf{w_1}),\dots,(0,\mathbf{w_n})\}$ and $\{(1,\mathbf{v_1}),\dots (1,\mathbf{v_m})\}$.

The cross-match statistic C, is a function of the word vectors $D = \{\mathbf{w_1,\dots,w_n,v_1,\dots,v_n}\}$ and the group labels $G = \{0,\dots,0,1,\dots,1\}$. If $H_0: W=V$ is true, then all the word vectors are i.i.d. ``sampled" from $W$ and the group labels are meaningless. It's as if the 0's and 1's were randomly assigned.

The cross-match test is performed as follows. For notational convenience ignore the group labels and treat the data as one sample $\{\mathbf{z_1,\dots,z_{n+m}}\}$ of size $n+m = N$ (assume for simplicity that $N$ is even). We define a $N \times N$ symmetric distance matrix, with row $k$ and column $l$ giving the distance (any distance metric can be used) between $\mathbf{z_k}$ and $\mathbf{z_l}$. Compute the optimal non-bipartite matching of the $\mathbf{z's}$ (match the vectors into non-overlapping pairs) that minimizes the total distances between the points in each pair.

Formally, we find a permutation $\hat{\sigma}$ of $\{1,\dots,N\}$ that minimizes $$Match(\sigma) = \sum_{i=1}^{N} d(Z_i,Z_{\sigma(i)})$$ where $i \neq \sigma(i)$ and $d$ is our chosen distance measure. The cross-match statistic $C$, is defined as the number of pairs that have group labels (0,1) or (1,0), the test rejects for small values of $C$.

If there is an odd number of word embedding vectors, then a psuedo-vector is added to the distance matrix at zero distance from everyone else. $\frac{N}{2}$ pairs are formed as before, and the pair containing the psuedo-vector is discarded (thus the least matchable word vector is discarded).

\subsection{Null Distribution of the Cross-Match Statistic}
One advantage of the cross-match test is that we can compute the exact distribution of the statistic $C$ under the null hypothesis $H_0$. Given $\frac{N}{2}$ paired vectors, let $c_0$ denote the observed number of the pairs with group labels (0,0), let $c_1$ denote the observed number of pairs with group labels (0,1) or (1,0) (this is our observed cross-match statistic) and finally let $c_2$ denote the observed number of pairs with group labels (1,1). The null distribution of $C$ in closed form is:

$$ f(c_1) = P(C=c_1) = \frac{2^{c_1}n!}{{{N}\choose{n}}c_0!c_1!c_2!} $$

where $\frac{N}{2} = c_0 + c_1 + c_2$. Having the null distribution in closed form also allows us to compute the exact p-value for our observed cross-match statistic. The resulting p-value is equal to $F(c_1)$ where $$ F(c_1) = P(C \leq c_1) = \sum_{c_1'=0}^{c_1} f(c_1') $$

A low p-value would suggests that we have evidence to reject the null hypothesis (at a given level of significance) that the word embedding vectors were ``sampled" from the same distribution. 

\section{Experiments}
In the following experiments, we demonstrate two different illustrative examples of the cross-match test. Our objective is to show the effectiveness of cross-match as a general tool for intrinsic evaluation of word embedding vectors.

\subsection{Embedding Similarity}
A bridge language (also referred to as a pivot language), is an artificial or natural language used as an intermediary for translation between two different languages. In machine translation, a bridge language is useful in low-resource situations where a good parallel corpora is not available for the target language. In such cases, a resource rich, linguistically similar language is used as a proxy in order to perform the required NLP task. For example in \citet{Tsvetkov2015} the authors use Arabic, Italian and French as bridge languages to perform Swahili-English, Maltese-English and Romanian-English translations respectively.

Assessing whether languages are linguistically similar is a reasonably difficult task and depends on the notion of similarity one uses (lexical, morphological etc.) In this experiment, we use cross-match to provide a quantitative measure to assess linguistic similarity between languages.

We use pre-trained word vectors (trained on Wikipedia using the skip-gram model in \citet{Bojanowski2016}) from Facebook's fastText library for several languages and calculate the cross-match statistic for several language pairs. Specifically, we randomly select 100,000 word vectors for each language (with the exception of Maltese and Swahili which have only 26,000 and 52,000 vectors respectively). Then for each language pair, we randomly sample 200 vectors and calculate the number of cross-matches between them using R's crossmatch package (\url{https://github.com/cran/crossmatch}). We repeat this 500 times for each language pair and report the average cross-match statistic.  

\begin{table}[h]
\begin{center}
\begin{tabular}{l|r}
\hline \bf Language Pair & \bf Cross-Match\\ \hline
English-French & 23.76 \\
English-Italian & 25.04 \\
English-Spanish & 23.36 \\
English-Portuguese & 18.44 \\
English-Arabic & 19.34\\
English-Maltese & 7.84\\
English-Romanian & 16.56 \\
English-Swahili & 17 \\
\hline
\end{tabular}
\end{center}
\caption{\label{table1} fastText vectors cross-match statistics for English-pair languages}
\end{table}

\begin{table}[h]
\begin{center}
\begin{tabular}{l|r}
\hline \bf Language Pair & \bf Cross-Match\\ \hline
Maltese-English & 7.84 \\
Maltese-French & 7.28 \\
Maltese-Italian & 9.20 \\
Maltese-Spanish & 6.76 \\
Maltese-Portuguese & 4.84 \\
Maltese-Arabic & 6.68 \\
Maltese-Romanian & 6.96 \\
Maltese-Swahili & 4.44 \\
\hline
\end{tabular}
\end{center}
\caption{\label{table2} fastText vectors cross-match statistics for Maltese-pair languages}
\end{table}

Tables \ref{table1} and \ref{table2} present the results of calculating the average number of cross-matches between several English-pair and Maltese-pair languages. We note that with a sample of 400 vectors (200 from each language) the maximum possible number of cross-matches is 200. Given that are our reported statistics are considerably lower than 200 we can safely conclude that the distributions from which the word embedding vectors were generated are different for different languages. In table \ref{table1} we note that the number of cross-matches between English and other romance languages (French, Italian, Spanish, Portuguese, Romanian) is noticeably higher than that between English and non-romance languages (Arabic, Maltese, Swahili). This corresponds with our notions of linguistic similarity between the languages, we certainly expect English to be more ``similar" to French than to Maltese. We also note that in table \ref{table2}, the Maltese-Italian pair has the highest cross-match statistic, thus supporting the choice of Italian as a bridge language for Maltese.

\subsection{Embedding Evaluation}
In this experiment, we use cross-match to assess the statistical significance of word embedding models. Despite the popularity of various different embedding models \cite{Mikolov2013a, Mikolov2013b, Pennington2014} it is not always clear whether one model represents a statistically significant improvement to other existing models (it maybe that all of them capture largely similar features of the text).

We consider four popular word embedding models: word2vec Skip-gram, word2vec CBOW, Glove and fastText all trained on the same English wikipedia corpus. Once again we take samples of size 200 from each method, caluclate the p-value between two pairs of methods using cross-match and then report the average p-value across 500 repeated iterations. 

\begin{table}[ht]
\small
\begin{center}
\begin{tabular}{l|l|l|l|l}
\hline & Skip & CBOW & Glove & FastText\\ \hline
Skip & - & 4.93e-26 & 2.39e-27 & 1.66e-23\\
CBOW & 4.93e-26 & - & 9.42e-25 & 2.71e-22  \\
Glove & 2.39e-27 & 9.42e-25 & - & 1.13e-23 \\
fastText & 1.66e-23 & 2.71e-22 & 1.13e-23 & - \\
\hline
\end{tabular}
\end{center}
\caption{\label{table3} p-values calculated using Cross-match}
\end{table}

The results in \ref{table3} show low p-values across all pairs of word embedding methods thus suggesting that they all seem to capture different aspects of the corpus they are modeling. In other words, using cross-match we have evidence to reject the null hypothesis that the vectors derived from any pair of models come from the same word embedding distribution. 

%We use the cross-match test on all model pairs and derive p-values for each pair. We expect to observe low p-values across all pairs suggesting for example that the vectors from Skip-gram are significantly different from Glove in terms of distributional similarity. In other words, using cross-match we're able to reject the null hypothesis that the vectors derived from both models come from the same word embedding distribution.

%The idea here is that one can use the cross match test to figure out whether the new proposed embedding model is a statistically significant improvement to the previous existing model. Take the vectors from Model 1 and the vectors from Model 2, run cross match and if the calculated p-value is below 0.05 you can reject the null hypothesis that distributions are the "same".

Lastly, we note that there are at present some computational constraints in performing the cross-match test. There exists a bottleneck in the calculation of the optimal non-bipartite matching and this makes performing the test for larger sample sizes currently intractable. However, we feel confident that this software issue can be easily overcome by writing custom routines (as opposed to using existing open-source code) and parallelizing the problem. As a result of our limited sample size, we not that it is possible that the power of our hypothesis test is low and thus we may be making type I errors (falsely rejecting the null). Nonetheless our initial results seem promising and are in line with our expectations.

\section{Conclusion}

In this work we introduced the cross-match test, an exact, distribution free, high-dimensional hypothesis test as an intrinsic evaluation metric for word embeddings. We were able to demonstrate on two illustrative examples that the test performs reasonably in line with our expectations and can potentially be a useful tool in assessing bridge languages for machine translation. Despite the initially promising results, much further work remains to be done in order to confirm the efficacy of cross-match in the context of word embeddings.

We posit that our main contribution is the introduction of the hypothesis testing framework as a method for intrinsic evaluation of vector representations. We observe that there is nothing notable about word embeddings or the cross-match test and our experiments could be extended for other vector representations (sentence, phrase etc.) using other modern two-sample hypothesis tests such as the popular maximum mean discrepancy \cite{Gretton2012}. Given the rich literature on hypothesis testing in statistics, there is certainly much to be explored here.

For future work we aim to focus solely on the problem of bridge languages in machine translation. Our objective is to conduct a larger scale study that is able to definitively show a strong correlation between the results of a hypothesis test on word embedding vectors, and their subsequent performance on the downstream machine translation task.

\section*{Acknowledgments}
We thank Ndapa Nakashole for several useful discussions helping formulate the problem and the anonymous reviewers for their feedback.

\bibliography{cse291}
\bibliographystyle{emnlp_natbib}

\end{document}